\documentclass{article}

\usepackage[final, nonatbib]{nips_2016}

\usepackage[utf8]{inputenc} 
\usepackage[T1]{fontenc}    
\usepackage{hyperref}       
\usepackage{url}            
\usepackage{booktabs}       
\usepackage{amsfonts}       
\usepackage{nicefrac}       
\usepackage{microtype}      
\usepackage{verbatim}       

\usepackage{subcaption}
\usepackage{graphicx}
\usepackage{float}
\usepackage{color,soul}
\usepackage{adjustbox}
\usepackage{amsmath}
\usepackage{bm}
\usepackage{mathtools}
\usepackage{algorithmic}
\usepackage{algorithm}
\usepackage{placeins}
\usepackage{tikz}
\usepackage{dsfont}
\usepackage{enumitem}
\usetikzlibrary{positioning}

\mathchardef\mhyphen="2D 
\DeclareMathOperator{\xentropy}{H}

\DeclareMathOperator*{\argmax}{\arg\max}
\DeclarePairedDelimiterX{\infdiv}[2]{(}{)}{%
	#1\;\delimsize\|\;#2%
}

\begin{document}

\title{Adversarial Images for Variational Autoencoders}

\author{
   Pedro Tabacof, Julia Tavares and Eduardo Valle\\
   RECOD Lab. — DCA / School of Electrical and Computer Engineering (FEEC)\\
   University of Campinas (Unicamp)\\
   Campinas, SP, Brazil \\
   \texttt{ \{tabacof, juliaat, dovalle\}@dca.fee.unicamp.br } \\
}

\maketitle
\begin{abstract}
We investigate adversarial attacks for autoencoders. We propose a procedure that distorts the input image to mislead the autoencoder in reconstructing a completely different target image. We attack the internal latent representations, attempting to make the adversarial input produce an internal representation as similar as possible as the target's.  We find that autoencoders are much more robust to the attack than classifiers: while some examples have tolerably small input distortion, and reasonable similarity to the target image, there is a quasi-linear trade-off between those aims. We report results on MNIST and SVHN datasets, and also test regular deterministic autoencoders, reaching similar conclusions in all cases. Finally, we show that the usual adversarial attack for classifiers, while being much easier, also presents a direct proportion between distortion on the input, and misdirection on the output. That proportionality however is hidden by the normalization of the output, which maps a linear layer into non-linear probabilities. 
\end{abstract}

\section{Introduction}

Adversarial attacks expressly optimize the input to ``fool'' models, e.g., in image classification, the adversarial input --- while visually tantamount to an ordinary original image --- leads to mislabelling with high confidence.

Here, we explore adversarial images for autoencoders — models optimized to reconstruct their inputs from compact internal representations. In an autoencoder, the attack targets not a single label, but a whole reconstruction. Our contributions include: 
\begin{itemize}[leftmargin=*] 
\item An adversarial attack on variational --- and, for comparison, deterministic --- autoencoders. Our attack aims not only at disturbing the reconstruction, but at fooling the autoencoder into reconstructing a completely different target image;
\item A comparison between attacks for autoencoders and for classifiers, showing that while the former is much harder, in both cases the amount of distortion on the input is proportional to the amount of misdirection on the output. For classifiers, however, such proportionality is hidden by the normalization of the output, which maps a linear layer into non-linear probabilities.
\end{itemize}

Evaluating generative models is hard~\cite{theis2015note}, there are no clear-cut success criteria for autoencoder reconstruction, and therefore, neither for the attack. We attempt to bypass that difficulty by analyzing how inputs and outputs differ across varying regularization constants.

The seminal article of Szegedy \emph{et al.} \cite{szegedy2013intriguing} introduced adversarial images, showing how to force a deep network to misclassify an image by applying nearly imperceptible distortions. Goodfellow \emph{et al.} \cite{goodfellow2014explaining} exploited the linear nature of deep convolutional networks to both attempt explaining how adversarial samples arise, and to propose a much faster technique to create them. Tabacof and Valle \cite{tabacof2015exploring} explored the geometry of adversarial regions, showing that they appear in relatively dense regions of the input space, and that shallow, simple classifiers tend to be more robust to them.

The existence of adversarial images lead to interesting questions on their significance, and even usefulness. Training models to resist adversarial attacks was advanced as a form of regularization~\cite{goodfellow2014explaining, miyato2015distributional}. Gu \emph{et al.} \cite{gu2014towards} used autoencoders to pre-process the input and try to reinforce the network against adversarial attacks, finding that although in some cases resistance improved, attacks with small distortions remained possible. A more recent trend is training adversarial \emph{models}, in which one attempts to generate “artificial” samples (from a generative model) and the other attempts to recognize those samples~\cite{goodfellow2014generative}. Makhzani \emph{et al.} \cite{makhzani2015adversarial} employ such scheme to train an autoencoder.

Although autoencoders appear in the literature of adversarial images as an attempt to obtain robustness to the attacks~\cite{gu2014towards}, and in the literature of adversarial training as models that can be trained with the technique~\cite{makhzani2015adversarial}, we are unaware of any attempts to create attacks targeted to them. In the closest related literature, Sara Sabour \emph{et al.} \cite{sabour2015adversarial} show that adversarial attacks can not only lead to mislabelling, but also manipulate the internal representations of the network. In this paper, we show that an analogous manipulation allows us to attack autoencoders, but that those remain much more resistant than classifiers to such attacks. 

\section{Autoencoders and Variational Autoencoders}

Autoencoders are models that map their input into a compact latent representation, and then, from such representation, build back the input (discounting some distortion). Therefore, autoencoders are trained to minimize the distortion between their input and their (reconstructed) output — plus regularization terms. The model comprises two parts: an encoder, which maps the input into the latent representation; and a decoder, which maps such representation into an output as close to the input as possible. In regular autoencoders, the training loss function may be as simple as the $\ell_2$-distance between input and output.

\begin{figure}[H]
    \centering
    \includegraphics[width=0.60\textwidth]{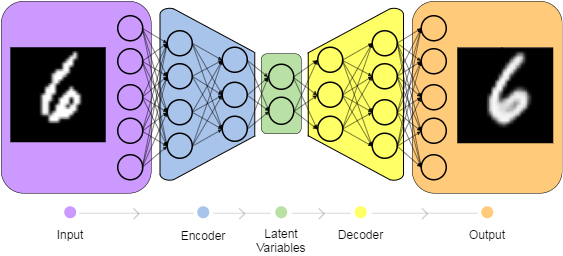}
    \caption{Autoencoders are models able to map their input into a (deterministic or stochastic) latent representation, and then to map such representation into an output similar to the input; those two maps form the two halves of the model: the encoder and the decoder.}
    \label{fig:autoencoderDiagram}
\end{figure}

Famous variants include sparse autoencoders, which use $\ell_1$-regularization~\cite{ng2011sparse}, and denoising autoencoders, which use implicit regularization by feeding noise to the input, while keeping the original input in the reconstruction loss term~\cite{vincent2010stacked}. An important offshoot are models with similar encoder–decoder structure, but which seek not to reconstruct the input, but to produce an output related to it (e.g., a segmentation map)~\cite{noh2015learning}.




A modern variant of growing popularity, variational autoencoders~\cite{kingma2013auto} interpret the latent representation through a Bayesian lens, thus offering a theoretical foundation for the reconstruction and regularization objectives. Variational autoencoders are probabilistic generative models, where we find the probability distribution of the data by marginalizing over the latent variables:

\begin{equation}
p_\theta(\bm{x}) = \int p_\theta(\bm{x},\bm{z}) d\bm{z} = \int p_\theta(\bm{x}|\bm{z})p(\bm{z})d\bm{z}
\end{equation} 

The likelihood $p_\theta(\bm{x}|\bm{z})$ is the probabilistic explanation of the observed data: in practice, often it is simply the output of the decoder network under a noise consideration (e.g. additive Gaussian noise for RGB pixels). The subscript $\theta$ comprises all decoder parameters, while $\bm{z}$ is the latent representation, over which we marginalize. The representation prior $p(\bm{z})$ is often the standard normal $\mathcal{N}(0,I)$~\cite{kingma2013auto}, but might be instead a discrete distribution (e.g. Bernoulli)~\cite{kingma2014semi}, or even some distribution with geometric interpretation (``what'' and ``where'' latent variables)~\cite{eslami2016attend}. Since the integration above is often intractable, we maximize its variational lower bound... 

\begin{equation}
\begin{aligned}
\mathrm{E}_{q_{\phi}(\bm{z}|\bm{x})}[\log p_\theta(\bm{x}|\bm{z})] - \mathrm{KL}\infdiv{q_\phi (\bm{z}|\bm{x})}{p(\bm{z})}
\ =\ -\mathrm{KL}\infdiv{q_\phi (\bm{z}|\bm{x})}{p(\bm{z}|\bm{x})}
\quad [\le\quad \log p(\bm{x})]  
\label{eq:vae}
\end{aligned}
\end{equation} 

...which is the Kullback--Leibler (KL) divergence between the approximate and the (unknown) exact posterior. Thus, maximizing the variational lower bound may also be interpreted as finding the best posterior approximation. In the context of variational autoencoders, such approximate posterior is usually an uncorrelated multivariate normal determined by the encoder network (with parameters $\phi$):

\begin{equation}
q_\phi(\bm{z}|\bm{x}) = \mathcal{N}(\bm{\mu}_{\phi}(\bm{x}), \exp(\bm{\sigma}_{\phi}^2(\bm{x})))
\label{eq:encoder}
\end{equation} 

We can approximate the likelihood expectation $\mathrm{E}_{q_{\phi}(\bm{z}|\bm{x})}[\log p_\theta(\bm{x}|\bm{z})]$ by Monte Carlo. As the prior and the approximated posterior are normal distributions, their KL divergence has analytic form~\cite{kingma2013auto}. We can use the reparameterization trick to reduce the variance of the gradient estimator~\cite{kingma2015variational}. 

The encoder and the decoder may be any neural network: a multilayer perceptron~\cite{kingma2013auto}, a convolutional network~\cite{radford2015unsupervised}, or even LSTMs. The latter are a recent development --- recurrent variational autoencoders --- which use soft attention to encode and decode patches from the input image~\cite{gregor2015draw, gregor2016towards}. 
Simulating a chain of samples from the latent variables and likelihood allows to denoise images, or to impute missing data (inpaint images)~\cite{rezende2014stochastic}. The latent variables of a variational autoencoder also allow visual analogy and interpolation~\cite{radford2015unsupervised}.



\section{Adversarial Images for Autoencoders}

Adversarial procedures minimize an adversarial loss to mislead the model (e.g., misclassification), while distorting the input as little as possible. If the attack is successful, humans should hardly be able to distinguish between the adversarial and the regular inputs~\cite{szegedy2013intriguing,tabacof2015exploring}. We can be even more strict, and only allow a distortion below the input quantization noise \cite{goodfellow2014explaining,sabour2015adversarial}.

To build adversarial images for classification, one can maximize the misdirection towards a certain wrong label \cite{szegedy2013intriguing, tabacof2015exploring} or away from the correct one \cite{goodfellow2014explaining}. The distortion can be minimized \cite{szegedy2013intriguing, tabacof2015exploring} or constrained to be small \cite{goodfellow2014explaining, sabour2015adversarial}. Finally, one often requires that images stay within their valid space (i.e., no pixels “below black or above white”). 

\begin{figure}[h]
    \centering
    \includegraphics[width=0.80\textwidth]{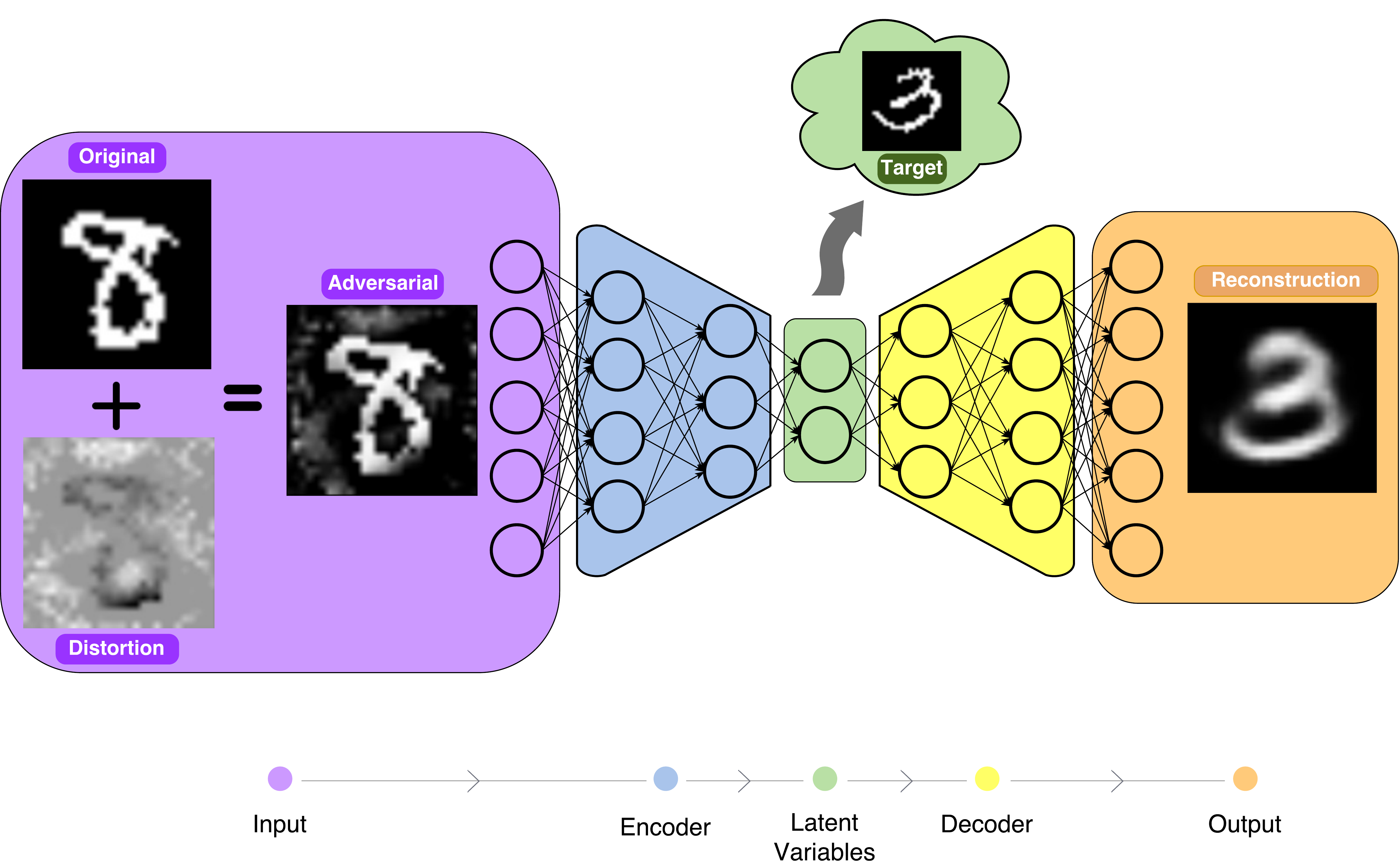}
    \caption{Adversarial attacks for autoencoders add (ideally small) distortions to the input, aiming at making the autoencoder reconstruct a different target. We attack the latent representation, attempting to match it to the target image's.}
    \label{fig:advAutoencoderDiagram}
\end{figure}

In autoencoders, there is not a single class output to misclassify, but instead a whole image output to scramble. The attack attempts to mislead the reconstruction: if a slightly altered image enters the autoencoder, but the reconstruction is wrecked, then the attack worked. A more dramatic attack --- the one we attempt in this paper --- would be to change slightly the input image and make the autoencoder reconstruct a completely different valid image (Fig.~\ref{fig:advAutoencoderDiagram}).

Our attack consists in selecting an original image and a target image, and then feeding the network the original image added to a small distortion, optimized to get an output as close to the target image as possible (Fig.~\ref{fig:advAutoencoderDiagram}). Our attempts to attack the output directly failed: minimizing its distance to the target only succeeded in blurring the reconstruction. As autoencoders reconstruct from the latent representation, we can attack it instead. The latent layer is the information bottleneck of the autoencoder, and thus particularly convenient to attack. We used the following adversarial optimization:

\begin{table}[h]
\begin{equation}
\label{eq:ae_adv}
\noindent
  \thinmuskip=\muexpr\thinmuskip*5/8\relax
  \medmuskip=\muexpr\medmuskip*5/8\relax
\begin{aligned}
& \underset{\bm{d}}{\min}
& & \Delta(\bm{z_{a}}, \bm{z_{t}}) + C\| \bm{d} \| \\
& \text{s.t.}
& & L \leq \bm{x} + \bm{d} \leq U \\
& & &  \bm{z_{a}} = \mathrm{encoder}(\bm{x} + \bm{d})
\end{aligned}
\end{equation} 
\end{table}

where $\bm{d}$ is the adversarial distortion; $\bm{z_{a}}$ and $\bm{z_{t}}$ are the latent representations, respectively, for the adversarial and the target images; $\bm{x}$ is the original image; $\bm{x}+\bm{d}$ is the adversarial image; 
$L$ and $U$ are the bounds on the input space; and $C$ is the regularizing constant the balances reaching the target and limiting the distortion.

We must choose a function $\Delta$ to compare representations. For regular autoencoders a simple $\ell_2$-distance sufficed; however, for variational autoencoders, the KL-divergence between the distributions induced by the latent variables not only worked better, but also offered a sounder justification. In our variational autoencoders, the $\bm{z_{*}}$ are uncorrelated multivariate normal distributions with parameters given by the encoder:

\begin{equation}
\mathrm{encoder}(\bm{x}) \sim \mathcal{N}(\bm{\mathrm{M}}_{\phi}(\bm{x}), \bm{\Sigma}_{\phi}(\bm{x}))
\end{equation}

where $\bm{\mathrm{M}}$ and $\Sigma$ are the representation mean vector, and (diagonal) covariance matrix output by the last layer of the encoder network; while $\phi$ are the autoencoder parameters --- learned previously by training it for its ordinary task of reconstruction. During the entire adversarial procedure, $\phi$ remains fixed.


\section{Data and Methods}

We worked on the binarized MNIST~\cite{lecun1998mnist} and SVHN datasets \cite{netzer2011reading}. The former allows for very fast experiments and very controlled conditions; the latter, while still allowing to manage a large number of experiments, provides much more noise and variability. Following literature~\cite{kingma2013auto}, we modeled pixel likelihoods as independent Bernoullis (for binary images), or as independent normals (for RGB images). We used Parmesan and Lasagne \cite{lasagne} for the implementation\footnote{The code for the experiments can be found at \url{https://github.com/tabacof/adv_vae}}.

The loss function to train the variational autoencoder (equation~\ref{eq:vae}) is the expectation of the likelihood under the approximated posterior plus the KL divergence between the approximated posterior and the prior. We approximate the expectation of the likelihood with one sample of the posterior. We extract the gradients of the lower bound using automatic differentiation and maximize it using stochastic gradient ascent via the ADAM algorithm \cite{kingma2014adam}. We used 20 and 100 latent variables for MNIST and SVHN, respectively. 
We parameterized the encoder and decoder as fully-connected networks in the MNIST case, and as convolutional and deconvolutional \cite{zeiler2010deconvolutional} networks in the SVHN case. After the training is done, we can use the autoencoder to reconstruct some image samples through the latent variables, which are the learned representation of the images.  An example of a pair of input image/reconstructed output appears in Fig.~\ref{fig:autoencoderDiagram}.

For classification tasks, the regularization term $C$ (Eq.~\ref{eq:ae_adv}) may be chosen by bisection as the smallest constant that still leads to success~\cite{tabacof2015exploring}. Autoencoders complicate such choice, for there is no longer a binary criterion for success. Goodfellow \emph{et al.} \cite{goodfellow2014explaining} and Sabour \emph{et al.}\cite{sabour2015adversarial} optimize differently, choosing for $\Delta$ an $\ell_\infty$-norm constrained to make the distortion imperceptible, while maximizing the misdirection. We found such solution too restrictive, leading to reconstructions visually too distinct from the target images. Our solution was instead to forgo a single choice for $C$, and analyze the behavior of the system throughout a series of values.

In our experiments, we pick at random 25 pairs of original/target images (axis “experiment” in  graphs). For each pair, we span 100 different values for the regularization constant $C$ in a logarithmic scale (from $2^{-20}$ to $2^{20}$), measuring the $\ell_2$-distance between the adversarial input and the original image (axis “distortion”), and the $\ell_2$-distance between the reconstructed output and the target image (axis “adversarial$-$target”). The “distortion” axis is normalized between 0.0 (no attack) and the $\ell_2$-distance between the original and target images in the pair (a large distortion that could reach the target directly). The “adversarial$-$target” is normalized between the $\ell_2$-distance of the reconstruction of the target and the target (the best expected attack) and the $\ell_2$-distance of the reconstruction of the original and the target (the worst expected attack). The geometry of such normalization is illustrated by the colored lines in the graphs of Fig.~\ref{fig:AdvExamples}. For variational autoencoders, the reconstruction is stochastic: therefore, each data point is sampled 100 times, and the average is reported.

For comparison purposes, we use the same protocol above to generate a range of adversarial images for the usual classification tasks on the same datasets. The aim is to contrast the behavior of adversarial attacks across the two tasks (autoencoding / classification). In those experiments we pick pairs of original image / adversarial class (axis “experiment”), and varying $C$ (from $2^{-10}$ to $2^{20}$), we measure the distortion as above, and the probability (with  corresponding logit) attributed to the adversarial (red lines) and to the original classes (blue lines). The axes here are no longer normalized, but we center at 0 in the “distortion” axis the transition point between attack failure and success --- the point where red and blue lines cross.



\section{Results and Discussion}

\begin{figure}[h!]
      \centering
      \begin{subfigure}[b]{0.45\textwidth}
          \includegraphics[width=\textwidth]{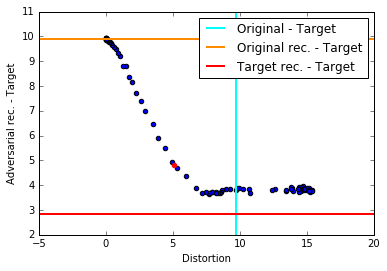}
          \label{fig:MNIST_VAE_Example_Dist}
      \end{subfigure}
      \begin{subfigure}[b]{0.45\textwidth}
          \includegraphics[width=\textwidth]{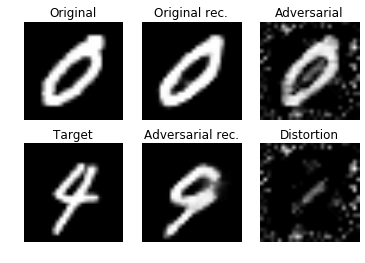}
          \label{fig:MNIST_VAE_Example}
      \end{subfigure}
      \begin{subfigure}[b]{0.45\textwidth}
          \includegraphics[width=\textwidth]{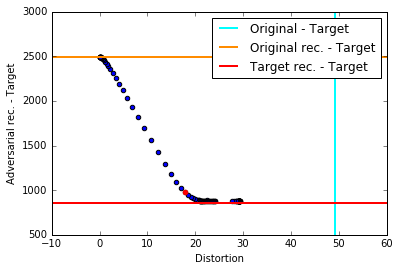}
          \label{fig:SVHN_VAE_Example_Dist}
      \end{subfigure}
      \begin{subfigure}[b]{0.45\textwidth}
          \includegraphics[width=\textwidth]{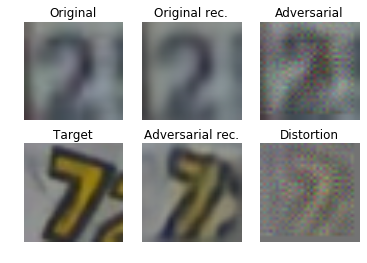}
          \label{fig:SVHN_VAE_Example}
      \end{subfigure}
      \caption{Top row: MNIST. Bottom row: SVHN. The figures on the left show the trade-off between the quality of adversarial attack and the adversarial distortion magnitude, with changing regularization parameter (implicit in the graphs, chosen from a logarithmic scale). The figures on the right correspond to the points shown in red in the graphs, illustrating adversarial images and reconstructions using fully-connected, and convolutional variational autoencoders (for MNIST and SVHN, respectively).}
      \label{fig:AdvExamples}
\end{figure}

We found that generating adversarial images for autoencoders is a much harder task than for classifiers. If we apply little distortion (comparable to those used for misleading classifiers), the reconstructions stay essentially untouched. To get reconstructions very close to the target's, we have to apply heavy distortions to the input. However, by hand-tuning the regularization parameter, it is possible to find trade-offs where the reconstruction approaches the target's and the adversarial image will still resemble the input (two examples in Fig. \ref{fig:AdvExamples}).

\begin{figure}[h]
      \centering
      \begin{subfigure}[b]{0.45\textwidth}
          \includegraphics[width=\textwidth]{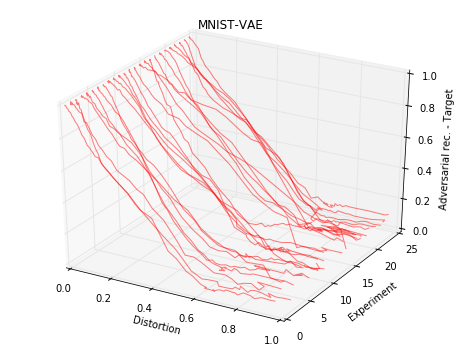}
          \label{fig:MNIST_VAE_3D}
      \end{subfigure}
      \begin{subfigure}[b]{0.45\textwidth}
          \includegraphics[width=\textwidth]{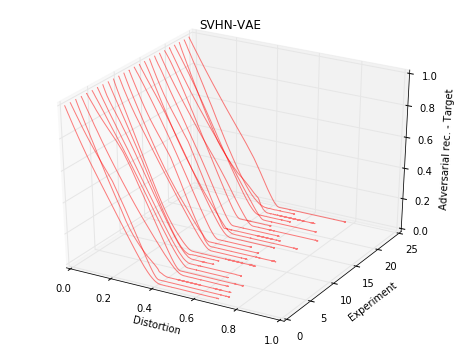}
          \label{fig:SVHN_VAE_3D}
      \end{subfigure}
      \begin{subfigure}[b]{0.45\textwidth}
          \includegraphics[width=\textwidth]{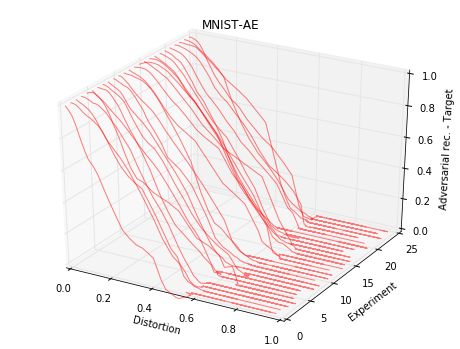}
          \label{fig:MNIST_AE_3D}
      \end{subfigure}
      \begin{subfigure}[b]{0.45\textwidth}
          \includegraphics[width=\textwidth]{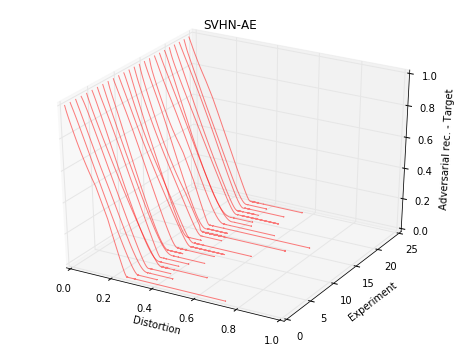}
          \label{fig:SVHN_AE_3D}
      \end{subfigure}
      \caption{Plots for the whole set of experiments in MNIST and SVHN. Top: variational autoencoders (VAE). Bottom: deterministic autoencoders (AE). Each line in a graph corresponds to one experiment with adversarial images from a single pair of original/target images, varying the regularization parameter $C$ (like shown in Fig.~\ref{fig:AdvExamples}). The “distortion” and “adversarial$-$target” axes show the trade-off between cost and success. The “hinge” where the lines saturate show the point where the reconstruction is essentially equal to the target's: the distortion at the hinge measures the resistance to the attack.
      }
\label{fig:Adv_3D}
\end{figure}

The plots for the full set of 25 original/target image pairs appear in Fig.~\ref{fig:Adv_3D}. All series saturate when the latent representation of the adversarial image essentially equals the target's. That saturation appears well before the upper distortion limit of 1.0, and provides a measure of how resistant the model is to the attack: Variational Autoencoders appear slightly more resistant than Deterministic Autoencoders, and MNIST much more resistant than SVHN. The latter is not surprising, since large complex models seem, in general, more susceptible to adversarial attacks.  Before the “hinge” where the attack saturates, there is a quasi-linear trade-off between input distortion and output similarity to target, for all combinations of dataset and autoencoder choice. We were initially hoping for a more non-linear behavior, with a sudden drop at some point in the scale, but data suggests that there is a give-and-take for attacking autoencoders: each gain in the attack requires a \emph{proportional} increase in distortion. 

\begin{figure}[h!]
     \centering
     \begin{subfigure}[b]{0.35\textwidth}
         \includegraphics[width=\textwidth]{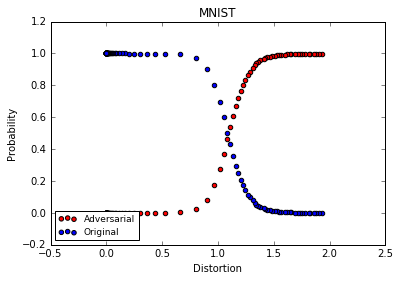}
         \label{fig:MNIST_clf_prob}
     \end{subfigure}
     \begin{subfigure}[b]{0.35\textwidth}
         \includegraphics[width=\textwidth]{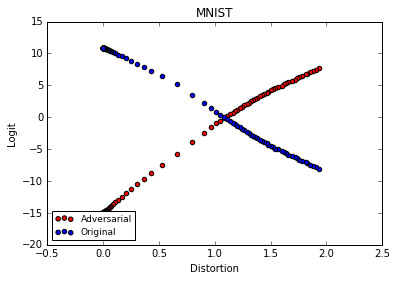}
         \label{fig:MNIST_clf_logit}
     \end{subfigure}
     \begin{subfigure}[b]{0.125\textwidth}
         \includegraphics[width=\textwidth]{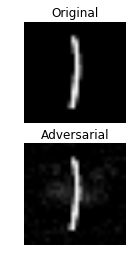}
         \label{fig:MNIST_clf_ex}
     \end{subfigure}
     \begin{subfigure}[b]{0.35\textwidth}
         \includegraphics[width=\textwidth]{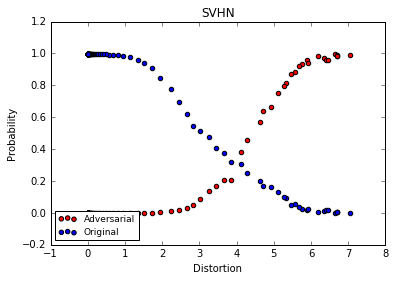}
         \label{fig:SVHN_clf_prob}
     \end{subfigure}
     \begin{subfigure}[b]{0.35\textwidth}
         \includegraphics[width=\textwidth]{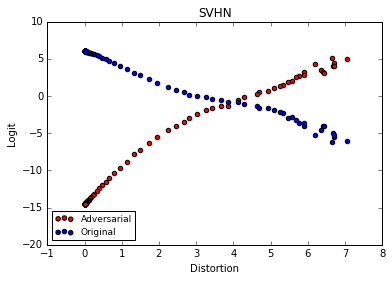}
         \label{fig:SVHN_clf_logit}
     \end{subfigure}
     \begin{subfigure}[b]{0.125\textwidth}
         \includegraphics[width=\textwidth]{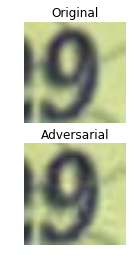}
         \label{fig:SVHN_clf_ex}
     \end{subfigure}
     \caption{Examples for the classification attacks. Top: MNIST. Bottom: SVHN. Left: probabilities. Middle: logit transform of probabilities. Right: images illustrating the intersection point of the curves. The adversarial class is ‘4’ for MNIST, and ‘0’ for SVHN. The red curve shows the probability/logit for the adversarial class, and the blue curve shows the same for the original class: the point where the curves cross is the transition point between failure and success of the attack.} 
     \label{fig:Adv_clf}
\end{figure}

\begin{figure}[h!]
	\centering
 	\begin{subfigure}[b]{0.45\textwidth}
 		\includegraphics[width=\textwidth]{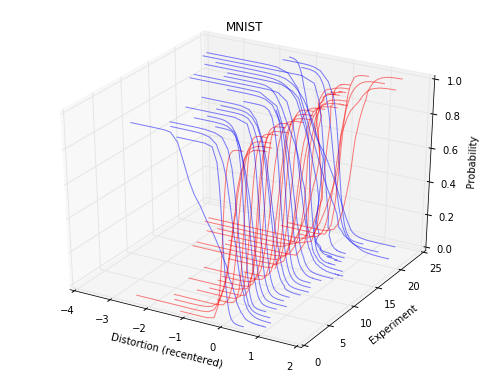}
 		\label{fig:MNIST_clf_prob_3D} 
     \end{subfigure}
	\begin{subfigure}[b]{0.45\textwidth}
		\includegraphics[width=\textwidth]{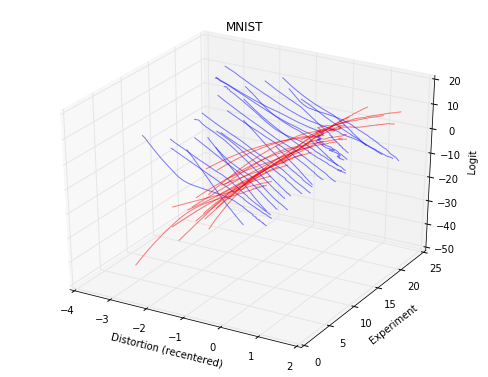}
		\label{fig:MNIST_clf_logit_3D} 
    \end{subfigure}
 	\begin{subfigure}[b]{0.45\textwidth}
 		\includegraphics[width=\textwidth]{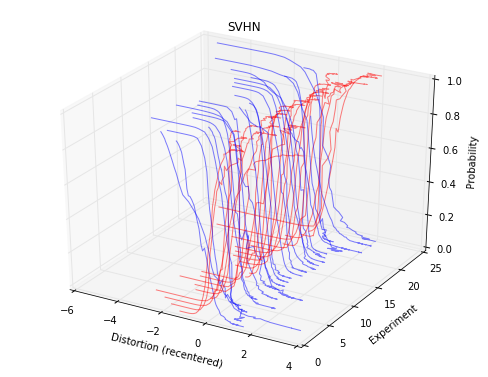}
 		\label{fig:SVHN_clf_prob_3D} 
     \end{subfigure}
	\begin{subfigure}[b]{0.45\textwidth}
		\includegraphics[width=\textwidth]{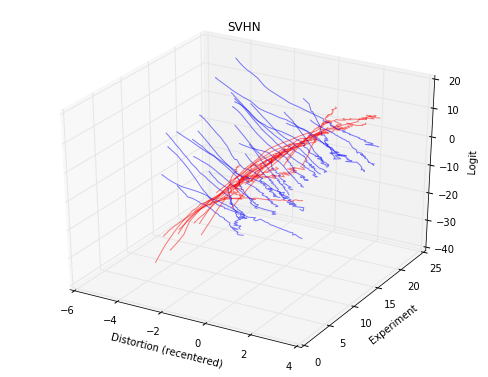}
		\label{fig:SVHN_clf_logit_3D} 
    \end{subfigure}
	\caption{Plot of whose set of experiments for classifiers. Top: MNIST. Bottom: SVHN. Left: probabilities. Right: logit transform of probabilities. Each experiment corresponds to one of the graphs shown in Fig.~\ref{fig:Adv_clf}, centered to make the crossing point between the red and blue lines stay at 0 in the “distortion” axis.}
	\label{fig:clf_3D}
\end{figure}

The comparison with the (much better-studied) attacks for classifiers, showed, at the beginning, a much different behavior: when we contrasted the probability attributed to the adversarial class \emph{vs.} the distortion imposed on the input, we observed the non-linear, sudden change we were expecting~(left column of Fig.~\ref{fig:clf_3D}). The question remained, however whether such non-linearity was intrinsic, or whether it was due to the highly non-linear nature of the probability scale. The answer appears in the right column of Fig.~\ref{fig:clf_3D}, where, with a logit transformation of the probabilities, the linear behavior appears again. It seems that the attack on classifiers show, internally, the same linear give-and-take present in autoencoders, but that the normalization of the outputs of the last layer into valid probabilities aids the attack: changes in input lead to proportional changes in logit, but to much larger changes in probability. That makes feasible for the attack on classifiers to find much better sweet spots than the attack on autoencoders (Fig.~\ref{fig:Adv_clf}). Goodfellow et al.~\cite{goodfellow2014explaining} suggested that the linearity of deep models make them susceptible to adversarial attacks. Our results seems to reinforce that such linearity plays indeed a critical role, with “internal” success of the attack being proportional to the distortion on inputs. On classification networks, however, which are essentially piecewise linear until the last layer, the non-linearity of the latter seems to compound the problem.


\section{Conclusion}

We proposed an adversarial method to attack autoencoders, and evaluated their robustness to such attacks. We showed that there is a linear trade-off between how much the adversarial input is similar to the original input, and how much the adversarial reconstruction is similar to the target reconstruction --- frustrating the hope that a small change in the input could lead to drastic changes in the reconstruction. 
Surprisingly, such linear trade-off also appears for adversarial attacks on classification networks, if we “undo” the non-linearity of the last layer. 
In the future, we intend to extend our empirical results to datasets with larger inputs and more complex networks (e.g. ImageNet) --- as well as to different autoencoder architectures. For example, the DRAW variational autoencoder \cite{gregor2015draw} uses feedback from the reconstruction error to improve the reconstruction --- and thus could be more robust to  attacks. We are also interested in advancing theoretical explanations to illuminate our results. 



\subsubsection*{Acknowledgments}

We thank Brazilian agencies CAPES, CNPq and FAPESP for financial support. We gratefully acknowledge the support of NVIDIA Corporation with the donation of the Tesla K40 GPU used for this research. Eduardo Valle is partially supported by a Google Awards LatAm 2016 grant, and by a CNPq PQ-2 grant (311486/2014-2).

\FloatBarrier 

\bibliographystyle{unsrt}
{\footnotesize
\bibliography{refs.bib}}

\end{document}